\documentclass[11pt]{article}

\usepackage[]{ACL2023}

\usepackage{times}
\usepackage{latexsym}

\usepackage[T1]{fontenc}
\usepackage[utf8]{inputenc}

\usepackage{microtype}

\usepackage{color}
\definecolor{my-color}{rgb}{0.28, 0.58, 0.28}

\usepackage{graphicx}
\usepackage{booktabs}
\usepackage{multirow}
\usepackage{adjustbox}
\usepackage{amssymb}
\usepackage{colortbl}
\usepackage{diagbox}
\usepackage{float}
\usepackage{amsmath}
\usepackage{algorithm} 
\usepackage{algpseudocode} 
\usepackage{scalerel,xparse}

\usepackage{pifont}
\usepackage{cuted}  % Provides the widetext environment
\usepackage{verbatim} % For the verbatim environment

\usepackage{listings}
\newcommand{\cmark}{\textcolor[RGB]{0,100,0}{\ding{51}}}

\newcommand{\xmark}{\textcolor{red}{\ding{55}}}

\definecolor{maroon}{cmyk}{0,0.87,0.68,0.32}

\definecolor{red1}{RGB}{253, 235, 232}
\definecolor{red2}{RGB}{243, 173, 172}
\definecolor{red3}{RGB}{228, 121, 121}

\definecolor{green1}{RGB}{209, 226, 255}
\definecolor{green2}{RGB}{163, 197, 255}
\definecolor{green3}{RGB}{121, 170, 255}

\definecolor{orange}{RGB}{255, 165, 102}
\definecolor{orange-fade}{RGB}{255,165,102}

\definecolor{orange-fade2}{RGB}{255,210,179}

\definecolor{sea-green}{RGB}{153, 226, 180}
\definecolor{sea-green-fade}{RGB}{194, 238, 210}

\definecolor{orangefade}{RGB}{255,230,204} % 稍淡橙色
\definecolor{seagreen}{RGB}{204,255,204}     % 稍淡绿色
\definecolor{seagreenfade}{RGB}{230,255,230} % 更淡绿色

\definecolor{Blue}{HTML}{2020df}
\definecolor{Pink}{HTML}{F08080}

\definecolor{purple}{RGB}{166,180,233}

\usepackage{arydshln}
\usepackage{hhline}

\usepackage{inconsolata}

\title{ReMedy: Learning Machine Translation Evaluation from \\ Human Preferences with Reward Modeling}

\author{Shaomu Tan \qquad Christof Monz\\
  Language Technology Lab\\ University of Amsterdam \\
  \texttt{\{s.tan, c.monz\}@uva.nl} \\}

\begin{document}

\maketitle
\begin{abstract}

%A key challenge in MT evaluation is the inherent noise and inconsistency of human ratings. Previous studies show that prompting and regressing Large Language Models excels in differentiating MT quality at the system level but struggle significantly at the segment level, especially for smaller models. In this work, we propose ReMedy, a novel MT metric framework that reformulates translation evaluation as a preference learning task. Instead of regressing on imperfect human ratings directly, ReMedy learns from relative translation quality using pairwise preference data, resulting in a more robust and reliable evaluation. Extensive experiments over WMT22-24, spanning 39 language pairs, 111 MT systems, and nearly one million segments, demonstrate that ReMedy achieves state-of-the-art performance on both segment- and system-level evaluation. Despite using only 9B parameters, ReMedy surpasses or reaches strong models such as GPT-4, PaLM-540B, XCOMET-Ensemble, and MetricX-13B. Further analysis demonstrates the robustness of ReMedy in handling translation error phenomena and diverse system quality levels, establishing preference learning as an effective paradigm for MT evaluation\footnote{We open source ReMedy models at \url{https://github.com/Smu-Tan/Remedy}.}.

A key challenge in MT evaluation is the inherent noise and inconsistency of human ratings. Regression-based neural metrics struggle with this noise, while prompting LLMs shows promise at system-level evaluation but performs poorly at segment level. In this work, we propose ReMedy, a novel MT metric framework that reformulates translation evaluation as a reward modeling task. Instead of regressing on imperfect human ratings directly, ReMedy learns relative translation quality using pairwise preference data, resulting in a more reliable evaluation. In extensive experiments across WMT22-24 shared tasks (39 language pairs, 111 MT systems), ReMedy achieves state-of-the-art performance at both segment- and system-level evaluation. Specifically, ReMedy-9B surpasses larger WMT winners and massive closed LLMs such as MetricX-13B, XCOMET-Ensemble, GEMBA-GPT-4, PaLM-540B, and finetuned PaLM2. Further analyses demonstrate that ReMedy delivers superior capability in detecting translation errors and evaluating low-quality translations.\footnote{We open source ReMedy models and all results at \url{https://github.com/Smu-Tan/Remedy}}

\end{abstract}

\section{Introduction}

\begin{figure}[h!]
    \centering
    \includegraphics[width=\linewidth]{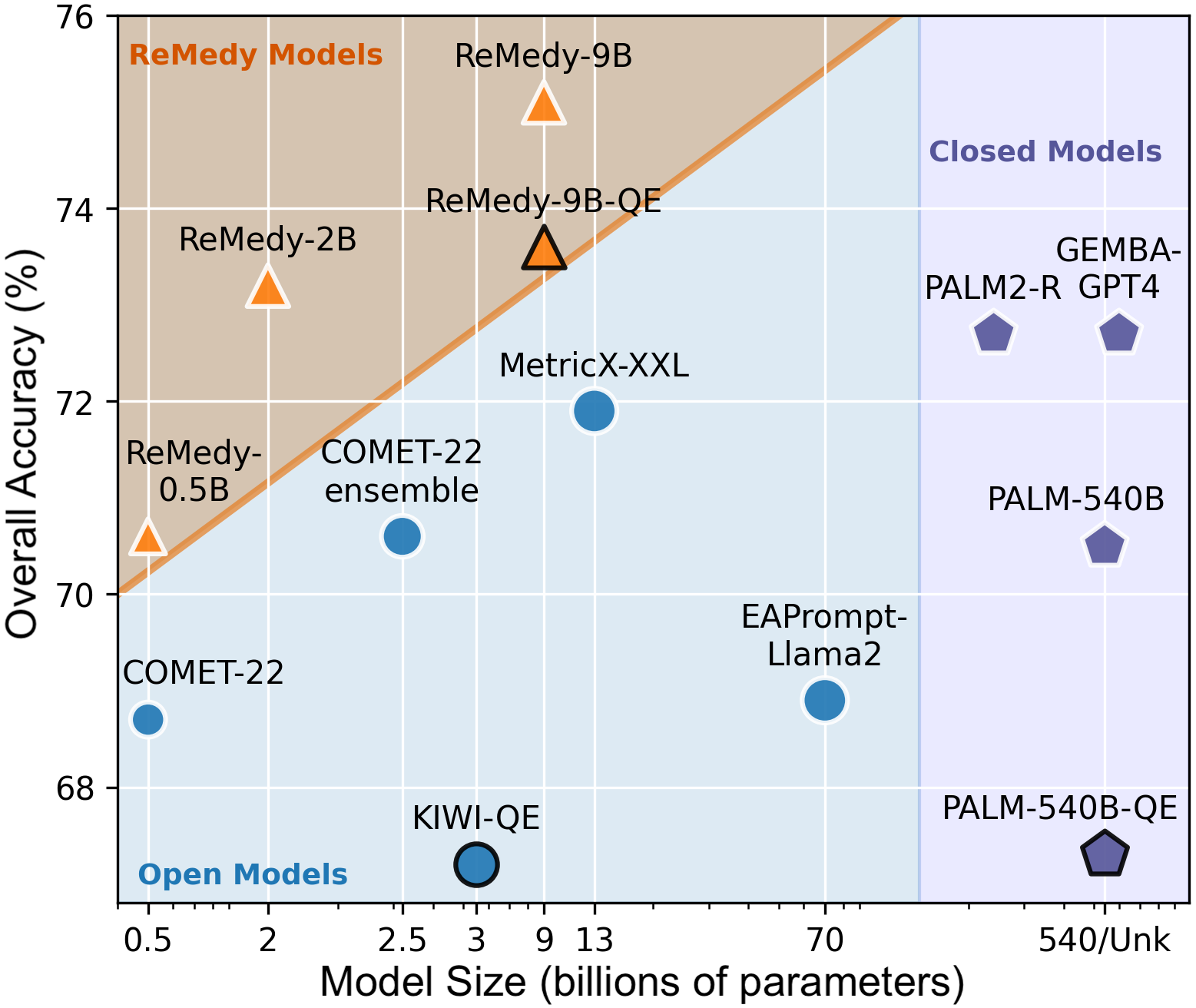}
%    \caption{Comparisons on WMT22 MQM set. We report the averaged accuracy over system-level pairwise accuracy and segment-level pairwise accuracy with tie calibration. For fair comparisons, our ReMedy models are trained on WMT17-20 metric data.}
    \caption{ 
    We report averaged accuracy over system- and segment-level pairwise accuracy for the WMT22 MQM set. The result shows that our largest ReMedy model achieves SOTA performance, surpassing previous WMT winners like MetricX-XXL, COMET, and massive fine-tuned closed LLMs like PaLM2.}
    \label{fig:teaser}
    %\vspace{-4mm}
\end{figure}

Machine Translation (MT) evaluation is crucial for benchmarking progress and guiding MT development. While string-based metrics like BLEU~\cite{papineni2002bleu,post2018call}, METEOR~\cite{banerjee2005meteor}, and ChrF~\cite{popovic2015chrf} have been widely used since 2002, they face persistent challenges: they poorly correlate with human judgments~\cite{freitag2022results}, struggle with reliability across diverse languages~\cite{goyal2022flores}, and fail to distinguish between translation systems of varying quality~\cite{przybocki2009nist}.

Neural metrics attempt to address these shortcomings. By leveraging pre-trained multilingual language models for regression tasks, they capture semantic equivalence beyond surface-level matching and extend language coverage~\cite{rei-etal-2020-comet,sellam-etal-2020-bleurt}. More recently, prompting Large Language Models (LLMs) for MT scoring has also shown promise in assessing translation quality across diverse contexts~\cite{fernandes2023devil,kocmi2023large}.

However, regression-based neural metrics have limitations. Human ratings are often noisy and inconsistent due to low inter-annotator agreement~\cite{rei2021references,song2025enhancing}, making direct regression unreliable. As a result, these models tend to be less robust in real-world scenarios, particularly when detecting translation error phenomena~\cite{amrhein-etal-2022-aces, moghe2025machine} and evaluating out-of-domain, low-quality systems compared to high-quality WMT submissions~\cite{lo2023metric,knowles2024mslc24}.

Recent work~\cite{kocmi2023large} also shows that prompting closed LLMs such as GPT-4 effectively differentiates translation quality at the system level, achieving SOTA correlations with human judgments. However, they perform substantially worse at the segment level---where individual translations are compared. This can be improved by extensive fine-tuning on MT evaluation data, yet massive LLMs like PaLM-2 still underperform much smaller models like MetricX-13B~\cite{juraska2024metricx}. Meanwhile, small, open LLMs continue to lag behind these closed LLMs~\cite{lu2024error,qian2024large,sindhujan2025llms}.

In this paper, we propose \textbf{\underline{Re}}ward \textbf{\underline{M}}odeling for \textbf{\underline{e}}valuating \textbf{\underline{d}}iverse translation qualit\textbf{\underline{y}} (\textbf{ReMedy}), a novel framework for MT evaluation that transforms pairwise human preferences into a robust reward signal. Unlike methods that regress over noisy absolute ratings or rely on pairwise classifiers that require quadratic comparisons, ReMedy learns from pairwise preferences, leading to more robust and reliable alignment with human judgments.

We conduct extensive experiments on the WMT22--24 metric shared tasks span 39 language pairs, 111 MT systems, and about 1 million testing segments. We show that using the same XLM-R-large foundation, ReMedy outperforms the regression-based COMET-22 model at both segment and system levels, matching the performance of the COMET-22 ensemble (5 models). Furthermore, our ReMedy-9B model surpasses larger models such as GPT-4, PaLM-540B, fine-tuned PaLM-2, and top WMT winners like xCOMET (24B ensemble) and MetricX (13B).

Analyses on the ACES~\cite{amrhein-etal-2022-aces,moghe2025machine} and MSLC~\cite{lo2023metric} challenge sets show that ReMedy is more robust in detecting translation errors across 146 diverse language pairs and evaluating low-quality translations, making it applicable to real-world MT deployment. Beyond standard evaluation tasks, we also explore how ReMedy can be integrated into Reinforcement Learning from Human Feedback (RLHF) pipelines, leveraging its robust preference-based framework to guide model updates for improved translation quality. Our key contributions are:

\paragraph{Reward Modeling for MT Assessment.} We introduce ReMedy, which leverages reward modeling for MT preferences to achieve better alignment with human judgment than regression approaches.
    
\paragraph{SOTA Performance with Fewer Parameters.} Our ReMedy-9B model achieves State-Of-The-Art results across WMT22--24 while requiring fewer parameters than previous methods (9B vs.\ 13B or 24B+ in ensemble models).

\paragraph{Enhanced Robustness in Challenging Scenarios.} ReMedy demonstrates superior performance in detecting translation error phenomena and reliably evaluates systems across a wide range of qualities.

\paragraph{ReMedy in MT-RLHF.} We show that replacing xCOMET with ReMedy in RLHF pipelines yields consistent performance gains, demonstrating its efficacy as a reward model for improving MT quality.

%Moreover, these frameworks demonstrate substantial performance degradation in reference-free scenarios, indicating continued dependence on parallel exemplars.

%(3) ReMedy's flexible design principles can accommodate various model architectures, providing adaptability for diverse multilingual modeling approaches. We summarize our key contributions as follows:

\section{Related Work}

Developing MT evaluation frameworks that align with human preference has remained challenging. %These preferences encompass subtle linguistic qualities and contextual appropriateness that are hard to quantify. 

\paragraph{String-Based Metrics.} Metrics like BLEU~\cite{papineni2002bleu} and ChrF~\cite{popovic2015chrf} rely on surface-level matching, which is computationally efficient but fails to capture semantic equivalence.

\paragraph{Learning MT Evaluation via Regression.} Recent approaches like COMET~\cite{rei-etal-2020-comet,rei2022comet}, xCOMET~\cite{guerreiro2024xcomet}, and MetricX~\cite{juraska2023metricx,juraska2024metricx} leverage pre-trained multilingual models such as XLM-R~\cite{conneau2020unsupervised} and mT5~\cite{xue2021mt5} to predict translation quality based on human-annotated assessments from WMT shared tasks. Despite improvements, these methods require large model size (\textgreater 10B) or ensembles (\textgreater 24B) for strong performance~\cite{freitag-etal-2024-llms}, often misclassify low-quality translations~\cite{lo2023metric}, and exhibit limited robustness against diverse error phenomena~\cite{amrhein-etal-2022-aces}.

\paragraph{LLM as Judge for MT Evaluation.} Alternatively, recent work has explored using LLMs as direct judges for MT evaluation. Closed models like GPT-4 and PaLM present competing system-level performance but struggle at the segment level~\cite{kocmi2023large}, even with extensive fine-tuning~\cite{fernandes2023devil}. Meanwhile, open LLMs perform much worse than closed ones~\cite{qian2024large} and present limitations in language inconsistency~\cite{sindhujan2025llms} and prompt design~\cite{lu2024error}.

\paragraph{Pairwise Quality Assessment (QE).} Early works~\cite{gamon2005sentence,sudoh2021translation} explored binary classification for MT assessment. More recently, MT-RANKER~\cite{moosa2024mt} revisits this by directly optimizing the logistic regression objective, enhanced with synthetic data. However, these approaches function as classifiers rather than a standalone metric. As a result, they cannot evaluate individual translations, requires quadratic comparisons for multiple systems, and operates solely as a Quality Assessment (QE) system without leveraging available references.

\section{ReMedy: Learning MT Metrics via Reward Modeling}

In this section, we introduce ReMedy, a novel MT metric framework that learns from human preferences. We first formalize the MT evaluation task and revisit regression methods, then we describe each component in ReMedy.

\subsection{Task Definitions}
Machine translation evaluation aims to assess the quality of translated text by assigning scores that correlate with human judgment. Formally, given a source sentence $\mathit{src}$, a candidate translation $\mathit{mt}$, and optionally a reference translation $\mathit{ref^*}$, an MT metric $M$ produces a quality score, as formalized in Eq~\ref{equation:mt_metric}. Here, $\mathit{ref^*}$ indicates that the reference is optional (reference-free when $\mathit{ref^*=\emptyset}$). Higher $M$ scores indicate better translation quality.

\begin{equation}\label{equation:mt_metric}
M(\mathit{src, mt, ref^*}) \rightarrow \mathbb{R}
\end{equation}

\subsection{Regression-based Approach} 
Recent neural MT metrics, such as COMET~\cite{rei-etal-2020-comet} and MetricX~\cite{juraska2023metricx}, are trained to predict human quality ratings $h$ by minimizing the Mean Squared Error (MSE) loss (Eq.~\ref{equation:regression}). However, human ratings suffer from inconsistencies and varying inter-annotator agreement. Prior work~\cite{rei2021references,song2025enhancing} has shown that inter-annotator agreement on high-quality WMT MQM datasets yields low to modest correlation, typically ranging from 0.2 to 0.45 Kendall-Tau correlation.

\begin{equation}\label{equation:regression}
\mathcal{L}_{\mathit{mse}} = \mathbb{E}_{(\mathit{src,mt,ref^*,h}) \in \mathcal{D}} [(M(\cdot) - h)^2]
\end{equation}

These inconsistencies pose challenges for regression approaches, as models struggle to learn stable patterns from inherently noisy data. To mitigate this, some MT metrics normalize human ratings using z-score transformations~\cite{rei2022comet,guerreiro2024xcomet}. However, ~\citet{juraska2023metricx, juraska2024metricx} found that while z-normalization improves segment-level performance, it can degrade system-level performance, highlighting the trade-offs inherent in regression-based methods. These shortcomings motivate our preference-based approach.

\subsection{ReMedy: Learn MT Metric with Pairwise Preference}

Recent advances in AI alignment have demonstrated the effectiveness of reward modeling for capturing human preferences~\cite{christiano2017deep} in areas such as helpfulness and safety~\cite{ouyang2022training, bai2022constitutional}. Inspired by these approaches, we propose ReMedy, an MT evaluation framework that learns to predict translation quality by 
modeling reward of pairwise human preferences rather than absolute scores.

\paragraph{Model Architecture.} 
ReMedy builds on a pretrained multilingual language model with the LM head removed and a linear scoring head added to produce a scalar quality score (reward $r$). For encoder-only models, the [CLS] hidden state is mapped to the score head. For decoder-only models, following~\citet{ouyang2022training} and \citet{touvron2023llama}, we use the hidden state of the final token as input to the linear head.

\paragraph{Preference Learning Framework.} 
Given a input $x=\{\mathit{src,ref^*}\}$, and two candidate translations $y^+=\mathit{mt^+}$ and $y^-=\mathit{mt^-}$, where human annotators prefer $\mathit{mt^+}$ over $\mathit{mt^-}$, our model learns a reward function $r_\theta(\mathit{x,y})$ that assigns higher scores to preferred translations. The model is trained with a pairwise ranking objective, combined with a reward regularization term.

\paragraph{Preference Ranking Loss.} 
The core of our method is a pairwise ranking loss based on the Bradley-Terry model~\cite{bradley1952rank,ouyang2022training}, which maximizes the probability of correctly ordering two translations according to human preference, as formalized in Eq.~\ref{equation:ranking}.

\begin{equation}\label{equation:ranking}
\mathcal{L_{\textit{bt}}} = -\log \sigma\bigl(r_\theta(x,y^+) - r_\theta(x,y^-) - m(r) \bigr)
\end{equation}

Here, the predicted reward scores for the preferred and non-preferred translations are denoted as $r_\theta(x,y^+)$ and $r_\theta(x,y^-)$, respectively. The margin $m(r)=h^+-h^-$ enforces a minimum separation between scores proportional to the difference in human ratings, ensuring the model's predictions align with the degree of human preference. $\sigma$ is the sigmoid function.

This Bradley-Terry loss models the probability that translation $\mathit{mt^+}$ is preferred over $\mathit{mt^-}$ as a function of their reward difference, encouraging the model to assign higher rewards to better translations with sufficient separation when margin $m(r)$ is integrated. In our experiments, we construct preference pairs using translations $\mathit{mt^+}$ and $\mathit{mt^-}$ with their raw human ratings $h^+$ and $h^-$, given the same source and reference input.

\paragraph{Reward Regularization.} We found that directly optimizing the ranking loss for MT evaluation leads to reward explosion---the model continuously increases scalar reward scores. This occurs because the ranking loss focuses on relative differences, allowing the model to grow rewards without bound. In addition, unlike helpfulness or safety reward modeling tasks~\cite{ouyang2022training}, where outputs often have large differences, translations typically differ only slightly (e.g., minor errors like omission or punctuation), and such small variations can cause the model to magnify reward discrepancies uncontrollably.

\begin{equation}\label{equation:reg}
\begin{aligned}
\mathcal{L}_{\textit{reg}} = \mathbb{E}_{r} [ \max&(r - \beta_{\text{upper}}, 0)^2 \\
&+ \max(\beta_{\text{lower}} - r, 0)^2 ]
\end{aligned}
\end{equation}

To stabilize training and ensure the reward function produces well-calibrated scores within a reasonable range, we apply a reward regularization term (Eq~\ref{equation:reg}). We set $\beta_{\text{upper}}=3$ and $\beta_{\text{lower}}=-3$, to penalizes rewards that exceed 3 or fall below -3, constraining outputs to an effective range that captures approximately 90\% of the sigmoid's variation. In section~\ref{sec:ablations}, we show that such regularization is crucial for preventing reward explosion during training, preventing degenerate performance where the model might inflate reward differences arbitrarily to satisfy the ranking objective.

\paragraph{Combined Objective.} Our final training objective combines the ranking and regularization losses (Eq~\ref{equation:final}), where $\lambda$ is a hyperparameter that controls the strength of regularization. We set $\lambda$ to 0.1, as higher values limit the ranking objective.

\begin{equation}\label{equation:final}
\mathcal{L_{\textit{final}}} = \mathcal{L_{\textit{bt}}} + \lambda \cdot \mathcal{L_{\textit{reg}}}
\end{equation}

\paragraph{Inference and Reward Calibration.}\label{sec:Calibration}
While ReMedy is trained with pairwise data, it can evaluate individual triplets $(\mathit{src}, \mathit{mt}, \mathit{ref^*})$ during inference to produce a scalar reward $r \in \mathbb{R}$. This avoids the quadratic comparisons of methods like MT-RANKER~\cite{moosa2024mt}.

Despite regularization during training, reward scores may exceed the bounds during inference in practice. To normalize rewards into the [0, 1] range and prevent clustering (which obscures quality differences), we calibrate \(r\) using an entropy-guided sigmoid function \( \sigma(r/\tau)\). The key idea is to find the optimal temperature \(\tau\) by maximizing the Shannon entropy across 20 bins, encouraging an even score spread in [0, 1]. Intuitively, this prevents scores from clustering in small regions (e.g., all good translations getting scores very close to 1.0).

\section{Experimental Setup}\label{sec:Experimental_Setup}

%We describe the experimental setups in this section. More details of the datasets are in Appendix~\ref{appendix:data}.

This section outlines our benchmark choices, baselines, and implementation details.

\subsection{Datasets and Benchmarks}

We selected three complementary benchmarks to evaluate MT quality from multiple perspectives.

\paragraph{WMT Metric Shared Tasks.} We use WMT22-24, standardized frameworks for comparing MT metrics. Following standard practice, we train on earlier data (from WMT17), validate on previous years, and test on the current year (See Appendix~\ref{appendix:data}). We use both high-quality Multidimensional Quality Metric (MQM) data and crowdsourced ratings like Direct Assessment (DA)\cite{bojar-etal-2017-results} and Scalar Quality Metrics (SQM)\cite{mathur-etal-2020-results}. 

For evaluation, we use: WMT22~\cite{freitag-etal-2022-results} (16 language pairs, 40 systems, 392,647 segments with MQM and DA+SQM annotations); WMT23~\cite{freitag-etal-2023-results} (11 language pairs, 29 systems, 282,926 segments); and WMT24~\cite{freitag-etal-2024-llms} (MQM subset: 3 language pairs, 32 systems, 68,502 segments; ESA subset: 9 language pairs, 40 systems, 232,289 segments).

\paragraph{ACES.} Translation Accuracy ChallengE Set (ACES)~\cite{amrhein-etal-2022-aces, moghe2025machine} covers 146 language pairs with 68 translation error phenomena grouped into 10 types. We use ACES to analyze a wide range of translation errors, from simple perturbations to complex discourse issues.

\paragraph{MSLC.} The Metric Score Landscape Challenge (MSLC)~\cite{lo2023metric} evaluates metrics on low and medium-quality translations in out-of-domain contexts, using transformer MT model checkpoints from various training stages to create a quality spectrum. We use MSLC to assess metrics' ability to distinguish low-to-medium quality translations.

\begin{table*}[h!]
\centering
\def\arraystretch{1.0} % 增加行间距
\resizebox{\linewidth}{!}{%
\begin{tabular}{llccccccc||c}
\toprule
\multirow{2}{*}{\textbf{Type}} & 
\multirow{2}{*}{\textbf{Methods}} & 
\multirow{2}{*}{$\boldsymbol{\theta}$} & 
\multirow{2}{*}{\textbf{ref?}} &
\multicolumn{1}{c}{\textbf{System-Level}} & 
\multicolumn{4}{c}{\textbf{Segment-Level} $\boldsymbol{\mathit{acc^*_{eq}}}$} &
\multicolumn{1}{c}{\textbf{Avg}} \\
\cmidrule(lr){5-5} \cmidrule(lr){6-9}
      & & & & \textbf{Acc} (3 LPs) & \textbf{Avg} & \textbf{En-De} & \textbf{En-Ru} & \textbf{Zh-En} & \textbf{Corr} \\
\midrule
\multirow{10}{*}{\begin{tabular}{c}\textbf{Closed}\\\textbf{Models}\end{tabular}} 
 & GEMBA-GPT4 (\textit{P})         & --   & \cmark & 89.8\%  & 55.6\% & 58.2\% & 55.0\% & 53.4\% & 72.7\% \\
 & PaLM (\textit{P})               & 540B & \cmark & 90.1\% & 50.8\% & 55.4\% & 48.6\% & 48.5\% & 70.5\% \\
 & PaLM-2 BISON (\textit{R})       & --   & \cmark & 88.0\%  & 57.3\% & 61.0\% & 51.5\% & 59.5\% & 72.7\% \\
 & PaLM-2 BISON (\textit{GC})      & --   & \cmark & 86.1\%  & 54.8\% & 59.2\% & 49.3\% & 56.0\% & 70.5\% \\
 & PaLM-2 UNICORN (\textit{R})     & --   & \cmark & 87.6\%  & 58.0\% & 61.1\% & 52.6\% & 60.4\% & 72.8\% \\
%\noalign{\vskip 2pt}
%\cdashline{2-10}[10pt/3pt]
%\noalign{\vskip 2pt}
 & PaLM (\textit{P})               & 540B & \xmark & 84.3\%  & 50.3\% & 56.1\% & 43.1\% & 51.8\% & 67.3\% \\
 & PaLM-2 BISON (\textit{R})       & --   & \xmark & 87.6\% & 57.5\% & 59.9\% & 53.4\% & 59.2\% & 72.6\% \\
 %& PaLM-2 BISON (\textit{GC})      & --   & \xmark & 86.1\%  & 53.2\% & 57.5\% & 47.3\% & 54.9\% & 69.7\% \\
 %& PaLM-2 UNICORN (\textit{GC})    & --   & \xmark & 86.1\%  & 52.9\% & 57.9\% & 45.6\% & 55.3\% & 69.5\% \\
\midrule
\multirow{5}{*}{\begin{tabular}{c}\textbf{Open}\\\textbf{Models}\end{tabular}} 
 & Llama2-EAPrompt (\textit{P})    & 70B  & \cmark & 85.4\%  & 52.3\% & 55.2\% & 51.4\% & 50.2\% & 68.9\% \\
 & COMET-22-DA (\textit{R})        & 0.5B & \cmark & 82.8\%  & 54.5\% & 58.2\% & 49.5\% & 55.7\% & 68.7\% \\
 & COMET-22 (\textit{R})           & ensemble & \cmark & 83.9\% & 57.3\% & 60.2\% & 54.1\% & 57.7\% & 70.6\% \\
 & MetricX-XXL (\textit{R})        & 13B  & \cmark & 85.0\%  & 58.8\% & 61.1\% & 54.6\% & 60.6\% & 71.9\% \\
 %& Llama2-EAPrompt (\textit{P})    & 70B  & \xmark & 85.8\%  & 52.0\% & 55.0\% & 51.6\% & 49.3\% & 68.9\% \\
 & COMETKiwi (\textit{R})          & ensemble & \xmark & 78.8\%  & 55.5\% & 58.3\% & 51.6\% & 56.5\% & 67.2\% \\
 
\cmidrule(lr){2-10}
\multirow{4}{*}{\begin{tabular}{c}\textbf{Ours}\end{tabular}} 

& \textnormal{ReMedy}$_{\textnormal{xlmr-22}}$ & 0.5B & \cmark & 85.8\% & 55.4\% & 58.3\% & 52.2\% & 55.8\% & 70.6\% \\

& \textnormal{ReMedy}$_\textnormal{2B-22}$ & 2B & \cmark & 90.5\% & 55.9\% & 58.0\% & 53.0\% & 56.6\% & 73.2\% \\

& \textnormal{ReMedy}$_\textnormal{9B-22}$ & 9B & \cmark & \textbf{91.2\%} & \textbf{58.9\%} & 61.0\% & 60.4\% & 55.4\% & \textbf{75.1\%} \\

& \textnormal{ReMedy}$_\textnormal{9B-22-QE}$ & 9B & \xmark & \underline{\underline{89.4\%}} & \underline{\underline{57.8\%}} & 59.4\% & 59.9\% & 54.2\% & \underline{\underline{73.6\%}} \\

\bottomrule 
\end{tabular}%
}
\caption{Evaluation on WMT22 MQM set. Following official WMT22 meta evaluations, we report system-level Pairwise Accuracy~\cite{kocmi2021ship} and segment-level pairwise accuracy with tie calibration~\cite{deutsch2023ties}, using Perm-Both statistical significance test~\cite{deutsch2021statistical}. \textit{P} denotes prompting; \textit{R} and \textit{GC} represent training with regression and generative classification objectives. \textbf{Bold} and \underline{underline} indicate the best metric and QE (no reference) models. COMET-22 and COMETKiwi are ensembled with 5x and 6x 0.5B models, respectively.}
\label{tab:wmt22}
\end{table*}

\subsection{Baselines}

    We compare ReMedy with strong closed LLMs and open WMT metric winners, covering both open metrics and commercial LLM approaches.

    \subsubsection{Closed Models}

    \paragraph{GEMBA (\textit{P}).} A zero-shot prompting (\textit{P}) approach using GPT-4~\cite{achiam2023gpt} for quality assessment~\cite{kocmi2023large}. 
    
    %While GEMBA achieves competitive system-level performance, it underperforms at the segment level compared to non-LLM metrics.

    \paragraph{PaLM (\textit{P}).} Like GEMBA,~\citet{fernandes2023devil} prompts PaLM-540B model~\cite{chowdhery2023palm} to generate translation quality scores.

    \paragraph{PaLM-2 Models.} ~\citet{fernandes2023devil} also fine-tuned PaLM-2 models using both Regression (\textit{R}) and Generative Classification (\textit{GC}) objectives with previous WMT data. They included BISON and UNICORN (second largest and largest in the PaLM-2 family, respectively) in the experoments.
    
    %Both BISON and UNICORN (second largest and largest in the PaLM-2 family, respectively), are used in the experoments.
    
    %The Regression models are trained on WMT15--20 DA data, while the GC model is trained on a combination of WMT20 DA and MQM data. Two sizes, BISON and UNICORN (second largest and largest in the PaLM-2 family, respectively), are evaluated.

    \subsubsection{Open Models}

    \paragraph{Llama2-EAPrompt (\textit{P}).} The Error Analysis Prompting~\cite{lu2024error} combines chain-of-thought reasoning with error analysis to score translations, emulating human evaluation. We report their strongest open model based on Llama2-70B.

    \paragraph{MetricX (\textit{R}).} A series of SOTA regression-based metrics from Google~\cite{juraska2023metricx,juraska2024metricx}, fine-tuned from mT5 models with two-stage fine-tuning and hybrid training recipes, augmented by synthetic data. We compare against the strongest MetricX variants for each year of the WMT sets.

    \paragraph{COMET (\textit{R}).} COMET methods utilize XLM-R pretrained encoders to model translation quality via sentence embeddings from the source, translation, and reference. For WMT22, we compare with COMET-22-DA (0.5B) and COMET-22-ensemble~\cite{rei2022comet}; for WMT23-24, we compare with XCOMET~\cite{guerreiro2024xcomet} (ensemble with 2 $\times$ 10.7B and 1 $\times$ 3.5B models).

\subsection{Implementation and meta-Evaluation}

We train ReMedy by fine-tuning two multilingual pre-trained foundation models: XLM-R~\cite{conneau2020unsupervised} and Gemma2~\cite{team2024gemma}, covering both encoder- and decoder-only types. We use XLM-R-Large (0.5B) and Gemma2 2B and 9B.

To ensure efficient training, we train our models using DeepSpeed~\cite{rajbhandari2020zero} in bfloat16 precision. We set the maximum sequence length to 512 tokens and use the Adam optimizer with a learning rate of 5e-6 and an effective batch size of 2048. We conduct experiments using 4 NVIDIA H100 GPUs, with VLLM~\cite{kwon2023efficient} for fast, efficient inference.

For meta evaluation, we adopt the official WMT Metric Share Task Toolkit.\footnote{MTME: \url{https://github.com/google-research/mt-metrics-eval}} Following the official setup, we report the Pairwise Accuracy (Acc) proposed by \citet{kocmi2021ship} at system-level results for WMT22-23, and Soft Pairwise Accuracy (SPA)~\cite{thompson2024improving,freitag-etal-2024-llms} for WMT24. For segment-level, we report pairwise accuracy with tie calibration ($\mathit{acc^*_{eq}}$)~\cite{deutsch2023ties} for all WMT22-24, with the Perm-Both statistical significance test~\cite{deutsch2021statistical}.

\section{Results and Analyses}

In this section, we analyze ReMedy's performance in correlating with human judgments. Our experiments show that ReMedy achieves SOTA results across WMT22-24 while maintaining parameter efficiency (sec~\ref{sec:results:corre}). Analyses on ACES and MSLC confirm that ReMedy reliably captures diverse translation errors and quality levels (sec~\ref{sec:results:analyses}). We also show that using ReMedy in RLHF pipelines leads to consistent performance gains (sec~\ref{sec:results:rlhf}).

\subsection{Correlation with Human Preference}\label{sec:results:corre}

We evaluate ReMedy on WMT22, WMT23, and WMT24, with detailed results provided in Tables~\ref{tab:wmt22}, \ref{tab:wmt23}, and \ref{tab:wmt24} (see Appendix~\ref{appendix:data} for additional details).

\subsubsection{ReMedy vs. Regression.}  
Table~\ref{tab:wmt22} shows that when fine-tuning the same XLM-R-Large (0.5B) foundation model, ReMedy outperforms the regression-based COMET-22-DA model by +2.6 points in system-level Acc and +0.9 in segment-level $\mathit{acc^*_{eq}}$ (verified by the Perm-Both statistical test). These results suggest that ReMedy delivers a more robust training signal than regression on noisy absolute ratings.

\subsubsection{ReMedy achieves SOTA results in WMT22-24.}

\textbf{WMT22}: Table~\ref{tab:wmt22} shows that while closed LLMs (e.g., PaLM-2) achieve high system-level accuracies, they often underperform at the segment level compared to open metrics like MetricX. Notably, ReMedy-0.5B reaches the overall performance of PaLM 540B with only 0.09\% of its parameters. Compared to the strongest fine-tuned PaLM-2 UNICORN (\textit{R}), ReMedy-2B exhibits a -2.1\% drop in segment-level $\mathit{acc^*_{eq}}$, yet ReMedy-2B still presents +0.4\% overall improvement. Lastly, ReMedy-9B achieves the best performance across both system and segment levels, surpassing the strongest PaLM-2 UNICORN (\textit{R}) model by +2.3 averaged score.

\textbf{WMT23}: As presented in Table~\ref{tab:wmt23}, ReMedy-9B outperforms winner models (XCOMET and MetricX-23) on all MQM and DA+SQM subsets including segment and system levels, with an average improvement of +1.9\% and +2.8\%. Furthermore, ReMedy-9B achieves these gains with significantly fewer parameters compared to the 13B MetricX-23 and ensemble XCOMET (totaling over 24B).

\textbf{WMT24}: Table~\ref{tab:wmt24} shows that ReMedy-9B achieves the highest segment-level accuracy ($\mathit{acc^*_{eq}}$), with a tiny decrease (-0.2\%) in system-level SPA. Overall, ReMedy-9B outperforms WMT24 winners by +1.2\% on average. While MetricX-24-Hybrid and XCOMET leverage synthetic data to enhance system-level performance, ReMedy is holds great promise to benefit from such augmentation in future work, as it requires only pairwise preference data rather than the absolute ratings needed by MetricX.

\begin{table}[h!]
\centering
\def\arraystretch{1.0}% 
\resizebox{\linewidth}{!}{%
\begin{tabular}{lrr|r}
\toprule
\multirow{2}{*}{\textbf{Methods}}& 
\multicolumn{1}{c}{\textbf{Sys}} & \multicolumn{1}{c|}{\textbf{Seg}} & \multicolumn{1}{c}{\textbf{Avg}}\\
\multirow{1}{*}{}& 
\multicolumn{1}{c}{\textbf{SPA}} & \multicolumn{1}{c|}{$\boldsymbol{\mathit{acc^*_{eq}}}$} & \multicolumn{1}{c}{\textbf{corr}}\\

 \midrule

GEMBA-ESA (\textit{P}) & 85.2\% & 57.6\% & 71.4\%  \\

MetricX-24-Hybrid (\textit{R}) & 85.6\% & 58.5\% & 72.1\% \\

XCOMET (\textit{R}) & \textbf{86.6}\% & 57.6\% & 72.1\% \\

\midrule

ReMedy$_\textnormal{9B-24}$ \textnormal{(Ours)} & 86.4\% & \textbf{60.2}\% & \textbf{73.3}\% \\

\bottomrule 
\end{tabular}%
}
\caption{Evaluation on WMT24 MQM set. We align with the official setup using SPA and $\mathit{acc^*_{eq}}$.}
\label{tab:wmt24}
\end{table}

\paragraph{Reference-Free ReMedy.} 
Although ReMedy is trained with reference, the reference-free ReMedy-QE achieves SOTA performance among all QE models in WMT22-24. Here, for the QE mode, the only difference is the reference sentence is none, which enables multiple modes for a single model.
%Although ReMedy is trained with reference, we found that without additional supervision like adaptation training or hybrid training, ReMedy achieves SOTA performance among all reference-free models in WMT22-24. Here, for the reference-free mode, the only difference is the reference sentence is none, which enables multiple modes for a single model.

\begin{table*}[h!]
\centering
\def\arraystretch{1.0}% 
\resizebox{\linewidth}{!}{%
\begin{tabular}{lcccccc||c}
\toprule
\multirow{2}{*}{\textbf{Method}}& 
\multirow{2}{*}{$\boldsymbol{\theta}$}& 
\multirow{2}{*}{\textbf{ref?} } & 
\multicolumn{2}{c}{\textbf{System-Level Acc}} & 
\multicolumn{2}{c}{\textbf{Segment-Level $\mathit{acc^*_{eq}}$}} &
\multicolumn{1}{c}{\textbf{Avg}}
\\

\cmidrule(lr){4-5} \cmidrule(lr){6-7}
      & 
      & 
      & MQM (3LPs)
      & SQM (8LPs) 
      & MQM (3LPs) 
      & SQM (8LPs) 
      & \textbf{Corr}\\
 \midrule

MetricX-23 (\textit{R}) & 13B & \cmark & 90.7\% & 86.3\% & 56.9\% & 57.0\% & 72.7\% \\
XCOMET (\textit{R}) & ensemble & \cmark & 92.8\% & 87.0\% & 57.7\% & 56.8\% & 73.6\% \\
GEMBA-MQM-GPT4 (\textit{P}) & - & \xmark & \textbf{94.5}\% & 89.9\% & 55.2\% & 38.0\% & 69.4\% \\
MetricX-23-QE (\textit{R}) & 13B & \xmark & 89.0\% & 87.0\% & 56.1\% & 56.7\% & 72.2\% \\
XCOMET-QE (\textit{R}) & ensemble & \xmark & 91.6\% & 87.1\% & 55.8\% & 55.2\% & 72.4\% \\
COMETKiwi (\textit{R}) & ensemble & \xmark & 91.1\% & 88.7\% & 54.6\% & 56.0\% & 72.6\% \\

\midrule

ReMedy$_\textnormal{9B-23}$ & 9B & \cmark & 94.1\% & \textbf{91.7\%} & \textbf{58.2}\% & \textbf{57.8}\% & \textbf{75.5}\% \\

ReMedy$_\textnormal{9B-23-QE}$ & 9B & \xmark & \underline{92.0\%} 
& \underline{91.7\%} & \underline{57.0\%} & \underline{56.8\%} & \underline{74.4\%} \\

\bottomrule 
\end{tabular}%
}
\caption{Evaluation on WMT23 Metric Shared task including MQM and DA+SQM (use SQM in table for simplicity) sets. \textbf{Bold} and \underline{underline} indicate the best metric and QE (no reference) models.}
\label{tab:wmt23}
\end{table*}

\subsection{Ablation Studies}\label{sec:ablations}

Table~\ref{tab:ablations} presents the ablation studies of ReMedy, using the Gemma2-2B as the foundation model.

\paragraph{Reward Explosion.} We first train vanilla ReMedy, a variant optimized solely with the Bradley-Terry loss, similar to most reward models~\cite{touvron2023llama,ouyang2022training}. During training, we observed that the model continuously increased the final scalar reward scores regardless of the input. This behavior is intuitive, as the Bradley-Terry loss optimizes only the reward differences. In this setup, the model learns that increasing all reward scores makes the sigmoid output larger, thereby reducing the training loss. As a result, it produces excessively high rewards (mean = 17.18, std = 5.37).

Adding Reward Regularization (+ reg.) effectively mitigates this reward explosion issue, stabilizing the reward distribution (mean = 1.33, std = 0.5) and improving average accuracy on the WMT22 MQM set by +7.0\%.

\begin{table}[h!]
\centering
\def\arraystretch{1.1}% 
\setlength{\tabcolsep}{4pt}
\resizebox{\linewidth}{!}{%
\begin{tabular}{lccccc}
\toprule
\textbf{Method} & 
\multicolumn{3}{c}{\textbf{MQM-22}} & 
\multicolumn{2}{c}{\textbf{Reward}}  \\
\cmidrule(lr){2-4} \cmidrule(lr){5-6}
      & \textbf{Sys}
      & \textbf{Seg}
      & \textbf{Avg}
      & \textbf{Mean} 
      & \textbf{Std} \\
 \midrule
%Vanilla-ReMedy    & 79.6\% & 52.2\% & 65.9\% & 17.18 & 5.37 \\
%+ reg.            & +11.3\% & +2.7\% & +7.0\% & 1.33  & 0.50 \\
%+ reg. + margin   & +10.2\% & +3.0\% & +6.6\% & 1.93  & 0.63 \\
%+ reg. + margin + cali & +10.9\% & +3.7\% & +7.3\% & 0.82 & 0.08 \\

Vanilla-ReMedy-2B    & 79.6\% & 52.2\% & 65.9\% & 17.18 & 5.37 \\
+ reg.            & 90.9\% & 54.9\% & 72.9\% & 1.33  & 0.50 \\
+ reg. + margin   & 89.8\% & 55.2\% & 72.5\% & 1.93  & 0.63 \\
+ reg. + margin + cali. & 90.5\% & 55.9\% & 73.2\% & 0.82 & 0.08 \\
\bottomrule 
\end{tabular}%
}
\caption{Ablation study of adding reward regularization (reg.), margin, and reward calibration (cali.) for ReMedy-2B on WMT22 test set.}
\label{tab:ablations}
\end{table}

%\paragraph{Margin and Inference Calibrations.} Furthermore, incorporating margin signals further enhances segment-level performance by +0.3, as margins provide additional information about the degree to which the preferred translation is favored over the rejected one. For the reward calibrations, we apply a sigmoid function that its temperature is guided by entropy (see section.~\ref{sec:Calibration}). Such transformation normalizes the rewards to [0,1] range while slightly enhancing the performance overall.

%\paragraph{Margin and Inference Calibrations.} Incorporating the rating difference as a margin signal enhances segment-level performance (+0.3 Acc) by informing the model about the degree of preference between translations. For the reward calibration, we apply a sigmoid function that its temperature is guided by entropy (see section.~\ref{sec:Calibration}). Such calibration normalizes the rewards to [0,1] range and ensures that subtle but meaningful differences in raw reward scores are preserved as distinguishable differences in the final [0, 1] scores. Applying this calibration slightly enhances the overall performance by +0.7\%. Notably, ReMedy achieves SOTA performance without calibration, which primarily enhances score interpretability. We provide more detailed analyses in Appendix~\ref{appendix:reward_calibration}.

\paragraph{Margin and Inference Calibrations.} Incorporating the rating difference as a margin signal enhances segment-level performance (+0.3 Acc) by informing the model about the degree of preference between translations. For reward calibration, we apply a sigmoid function with its temperature guided by entropy (see section~\ref{sec:Calibration}). This calibration normalizes rewards to the [0,1] range while preserving meaningful distinctions between translations of similar quality, slightly improving overall performance by +0.7\%. Notably, ReMedy achieves SOTA performance without calibration, which only serves for normalization purposes. We provide more detailed analyses in Appendix~\ref{appendix:reward_calibration}.

%. The key idea is to find the optimal sigmoid temperature (\(\tau\)) that spreads the calibrated scores most evenly across the entire [0, 1] interval, maximizing their distribution entropy. Intuitively, this prevents scores from clustering in small regions (e.g., all good translations getting scores very close to 1.0). By utilizing the full [0, 1] range effectively, the calibration ensures that subtle but meaningful differences in raw reward scores are preserved as distinguishable differences in the final [0, 1] scores. This improved separation between scores for translations of varying quality leads to a better reflection of the relative human judgments, thus directly improving correlation performance.

\begin{figure}[h!]
    \centering
    \includegraphics[width=\linewidth]{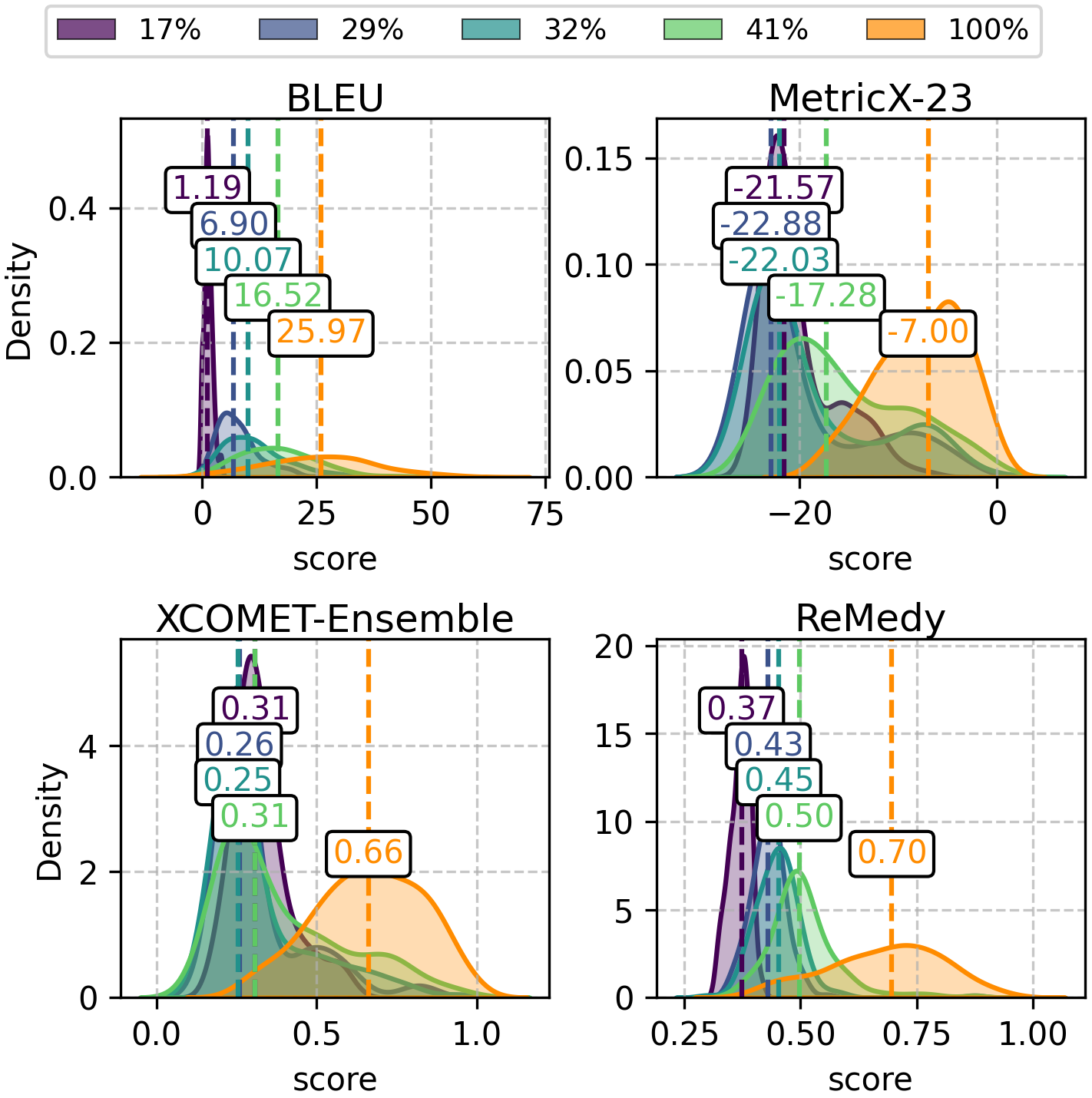}
    \caption{Kernel density plots of quality scores at various model checkpoints. Percentages indicate training progress stages, with dashed lines marking mean scores.}
    \label{fig:dist}
    %\vspace{-2mm}
\end{figure}

\subsection{Analyses on Challenge sets}\label{sec:results:analyses}

\begin{table*}[t]
\centering
\def\arraystretch{1.0}% 
\resizebox{\linewidth}{!}{%
\begin{tabular}{lcrrrrrrrrrr|r}
\toprule
& ref? & Add & Omi & Mis-T & Un-T & DNT & Over & Under & RW-K & WL & Punc & \textbf{ACES} \\
%\midrule
% \# Examples & 999 & 999 & 24457 & 1300 & 100 & 1000 & 1000 & 2948 & 2000 & 1673 & \\
\midrule
BLEU & \cmark & 0.75 & 0.44 & -0.23 & 0.36 & 0.60 & -0.84 & -0.86 & -0.77 & 0.66 & 0.64 & -2.8 \\
ChrF & \cmark & 0.64 & 0.78 & 0.16 & \textbf{0.78} & \textbf{0.96} & -0.70 & -0.59 & -0.29 & \textbf{0.69} & 0.74 & 3.71 \\

\midrule

%\rowcolor[gray]{0.9} COMET-20-QE* & \xmark & -0.538 & 0.397 & 0.378 & 0.135 & 0.120 & 0.622 & 0.442 & 0.322 & -0.505 & 0.251 & 6.61 \\

%COMET-20 & \cmark & 0.437 & 0.808 & 0.378 & 0.748 & 0.900 & 0.314 & 0.112 & 0.267 & 0.033 & 0.706 & 12.27 \\
MetricX-13B & \cmark & -0.10 & 0.53 & 0.58 & 0.65 & 0.88 & 0.75 & 0.55 & 0.71 & -0.32 & 0.37 & 13.54 \\
%BLEURT-20 & 0.437 & 0.810 & 0.429 & \underline{0.748} & 0.860 & 0.200 & 0.014 & 0.401 & 0.533 & 0.649 & 12.06 \\
COMET-22 & \cmark & 0.33 & 0.81 & 0.57 & 0.54 & 0.90 & 0.69 & 0.54 & 0.57 & -0.32 & 0.54 & 16.41 \\
KG-BERTScore & \cmark & \textbf{0.79} & 0.81 & 0.49 & -0.46 & 0.76 & 0.65 & 0.53 & 0.49 & 0.31 & 0.26 & 17.49 \\

%metricx\_xl\_DA\_2019 & 0.395 & 0.852 & 0.545 & 0.722 & 0.940 & 0.692 & 0.376 & 0.740 & 0.521 & 0.670 & 17.29 \\
%metricx\_xl\_MQM\_2020 & -0.281 & 0.670 & 0.523 & 0.579 & -0.740 & 0.718 & 0.602 & 0.705 & -0.126 & 0.445 & 13.10 \\
%metricx\_xxl\_DA\_2019 & 0.303 & 0.832 & 0.580 & 0.762 & 0.920 & 0.572 & 0.246 & 0.691 & 0.250 & 0.630 & 15.35 \\

COMET-Kiwi-22 & \xmark & 0.36 & 0.83 & 0.63 & 0.23 & 0.78 & 0.74 & 0.57 & 0.58 & -0.36 & 0.49 & 16.95 \\
MT-Ranker-13B & \xmark & 0.65 & \textbf{0.97} & 0.63 & 0.25 & 0.84 & 0.63 & 0.54 & 0.66 & -0.53 & \textbf{0.97} & 18.46 \\

\midrule

ReMedy$_\textnormal{2B-22}$ & \cmark & 0.35 & 0.72 & 0.66 & 0.63 & 0.70 & 0.79 & 0.55 & 0.82 & 0.20 & 0.64 & 17.74 \\
ReMedy$_\textnormal{9B-22}$ & \cmark & 0.49 & 0.86 & 0.71 & 0.70 & 0.76 & \textbf{0.81} & 0.56 & \textbf{0.89} & 0.31 & 0.60 & \textbf{19.90} \\

ReMedy$_\textnormal{2B-22-QE}$ & \xmark & 0.05 & 0.69 & 0.67 & 0.11 & 0.50 & 0.73 & 0.52 & 0.76 & -0.17 & 0.52 & 14.49 \\
ReMedy$_\textnormal{9B-22-QE}$ & \xmark & 0.48 & 0.81 & \textbf{0.73} & 0.39 & 0.56 & \textbf{0.81} & \textbf{0.59} & 0.87 & 0.04 & 0.58 & 18.93 \\

\bottomrule
\end{tabular}%
}
\caption{Average Kendall’s tau-like correlation results for the ten error categories spaning 68 translation phenomena. ACES-Score represents the overall performance across all categories (see details in appendix~\ref{appendix:aces}). Addition (Add), Omission (Omi), Mis-T (Mistranslation), Un-T (Untranslated), DNT (Do Not Translate), Over (Overtranslation), Under (Undertranslation), RW-K (Real-World Knowledge), WL (Wrong Language), Punc (Punctuation).}
\label{tab:ACES}
\end{table*}

In addition to the WMT benchmarks, we analyze ReMedy's performance in detecting translation errors and out-of-domain low-quality translations.

\paragraph{MSLC Challenge Set.}  
On the MSLC challenge set, ReMedy provides reliable quality scores across a wide range of translation outputs, effectively distinguishing between low- and medium-quality translations. As shown in Figure~\ref{fig:dist}, unlike XCOMET and MetricX, ReMedy presents a clear quality boundary for the English-German MT model for its different checkpoints, especially for out-of-domain low and medium quality (corresponding to 1 to 16 BLEU scores) translations.

\paragraph{ACES Challenge Set.}  

As shown in Table~\ref{tab:ACES}, ReMedy-9B achieves new SOTA results on the ACES benchmark that covers 146 language pairs, demonstrating the highest overall correlation (ACES score) with human judgments in detecting 68 diverse translation error phenomena. 

We noticed that all neural metrics perform poorly on the Wrong Language (WL) phenomenon, this is intuitive since such errors contain semantic equivalent but off-targeted~\cite{tan2023towards} translations. Incorporating synthetic data holds promise, and we leave them to future work.

\subsection{ReMedy in RLHF Pipelines}\label{sec:results:rlhf}

Lastly, we integrated ReMedy as a reward model in Reinforcement Learning from Human Feedback (RLHF) pipelines. We implement Contrastive Preference Optimization (CPO)~\cite{xucontrastive} based on the ALMA-13B~\cite{xuparadigm} model. Following the original CPO setup, we remain the training data unchanged, then use ReMedy-9B to score References, GPT-4 and ALMA translations. We then conduct CPO tuning on ALMA-13B with LoRA using the same hyper-parameter, then evaluate the final models with greedy decoding.

% due to computational constraints

\begin{table}[h!]
  \centering
  \def\arraystretch{1.1}% 
  \setlength{\tabcolsep}{3pt}
  \resizebox{\columnwidth}{!}{%
    \begin{tabular}{lccccc}
      \toprule
      \multicolumn{1}{l}{\textbf{RM}} &
      \multicolumn{1}{r}{\textbf{BLEU}} &
      \multicolumn{1}{r}{\textbf{COMET22}} &
      \multicolumn{1}{r}{\textbf{KIWI}} &
      \multicolumn{1}{r}{\textbf{XCOMET}} &
      \multicolumn{1}{r}{\textbf{ReMedy}}
      \\
      \midrule
      \rowcolor[gray]{0.9}
      \multicolumn{6}{c}{Results on WMT22 Testset (10 LPs)} \\
      XCOMET & 28.6 & 85.6\%  & 81.9\%  & 90.2\% & 80.8\% \\
      ReMedy & \textbf{29.8} & \textbf{85.9\%} & \textbf{82.3\%}  & \textbf{90.3\%} & \textbf{81.1\%} \\
      \rowcolor[gray]{0.9}
      \multicolumn{6}{c}{Results on WMT23 Testset (6 LPs)} \\
      XCOMET & 28.0 & 83.0\% & 76.9\% & 88.1\% & 80.6\% \\
      ReMedy & \textbf{29.4} & \textbf{83.3\%} & \textbf{77.1\%} & \textbf{88.2\%} & \textbf{81.1\%} \\
      \bottomrule
    \end{tabular}%
  }
  \caption{Performance of using XCOMET and ReMedy-9B as reward models for ALMA13B-CPO tuning on WMT22 and WMT23 general MT testsets. 
  }
  \label{tab:RLHF}
\end{table}

To avoid the metric interference~\cite{pombal2025adding}, i.e., use same metrics for both model tuning and evaluation, we report results based on various metrics including BLEU, COMETKIWI-10B, XCOMET, and ReMedy-9B. Table~\ref{tab:RLHF} shows that replacing the XCOMET reward model with ReMedy-9B yields consistent performance gains on all metric scores, underscoring ReMedy's versatility and potential for downstream MT improvements.

\section{Conclusions}
To address the challenges of noisy and inconsistent human ratings in MT evaluation, we introduced ReMedy, a novel framework leveraging reward modeling, augmented by reward regularization and calibration, to learn directly from pairwise human preferences. Our extensive experiments on WMT22-24 demonstrate that ReMedy achieves state-of-the-art performance at both segment and system levels. Notably, our 9B parameter ReMedy model surpasses significantly larger models, including GPT-4, PaLM-540B, XCOMET-Ensemble, and MetricX-13B. Further analyses confirmed its robustness on challenge sets designed to test error detection and handling of varying quality levels. Additionally, ReMedy's integration into RLHF pipelines highlights its potential as an effective reward model for improving MT systems. ReMedy shows that reward modeling with preference learning offers a more robust, efficient, and human-aligned approach to machine translation evaluation.

\section*{Limitations}
In this paper, we do not include the utilization of synthetic data in MT evaluation. Previous studies such as MetricX, XCOMET found constructing synthetic data for out-of-domain and fine-grained translation errors can improve the overall performance and form more robust systems. In this work, we focus more on how to improve the MT metric system with current available open-source data. However, ReMedy holds great promise in leveraging synthetic data, since it only requires pairwise preference data rather than absolute ratings like MetricX or XCOMET requires, we leave this to future work.

We noticed that for the WMT24 ESA subset, ReMedy performs slightly worse than MetricX and XCOMET (see appendix~\ref{appendix:results}). In specific, we found gaps mostly on English-Hindi and English-Icelandic pairs, where LLM-based approaches like GEMBA-MQM-GPT4 also present lower performance. We hypothesize this could be due to the nature of these language pairs remaining low-resource for pre-trained decoder-only LLMs. Nonetheless, we found that ReMedy-9B-22 outperforms MetricX and COMET on unseen extremely low-resource language pairs like English-Livonian, and Yakut-Russian in the WMT22 test set. We plan to look at the potential reasons in the future.

\section*{Broader Impact}
We acknowledge several ethical considerations in MT evaluation research. To address the risk of mistranslation, we prioritize high-quality data from WMT Metric Shared tasks, though fairness challenges persist as metrics may perform inconsistently across the linguistic spectrum, particularly for low-resource languages. Furthermore, MT systems and evaluation metrics can perpetuate societal biases present in training data, such as human biases.

\section*{Acknowledgments}
This research was funded in part by the Netherlands Organization for Scientific Research (NWO) under project number VI.C.192.080.

% Entries for the entire Anthology, followed by custom entries
\bibliography{anthology,custom}
\bibliographystyle{acl_natbib}

\newpage
\appendix

\section{Appendix}\label{sec:appendix}

\subsection{Data}\label{appendix:data}

\subsubsection{WMT Metric Shared Tasks}

WMT Metric Shared Tasks provided a standardized framework for comparing automatic MT evaluation metrics using human assessments since 2008~\cite{callison-burch-etal-2008-meta}.  In WMT22-24, various annotation methods have been employed. Among these, the Multidimensional Quality Metric (MQM) stands out due to its reliance on professional translators for fine-grained error annotations, making it particularly reliable for assessing high-quality MT outputs~\cite{freitag2021experts}. 

In contrast, other evaluation approaches including Direct Assessment (DA)~\cite{bojar-etal-2017-results}, Scalar Quality Metrics (SQM)~\cite{mathur-etal-2020-results}, Error Span Analysis (ESA)~\cite{freitag-etal-2024-llms} are based on crowdsourced ratings, which may not always capture the same level of nuance and precision~\cite{freitag2021experts}.

Human assessments in WMT22-24 include four types of annotations below. MQM is considered as the highest-quality assessment, which is more reliable for high-quality MT predictions~\cite{freitag2021experts}.

\begin{itemize}
    \item \textbf{Multidimensional Quality Metric (MQM)}: Professional translators provide fine-grained error annotations~\cite{freitag-etal-2021-experts}.
    \item \textbf{Direct Assessment (DA)}: Crowdsourced holistic quality ratings on a 0--100 scale~\cite{bojar-etal-2017-results}.
    \item \textbf{Scalar Quality Metrics (SQM)}~\cite{mathur-etal-2020-results}: A simplified version of MQM with fewer error categories.
    \item \textbf{Error Span Analysis (ESA)}~\cite{freitag-etal-2024-llms}: 0--100 Ratings accompanied by error span annotations.
\end{itemize}

Following standard practice~\cite{guerreiro2024xcomet}, we train on earlier data (e.g., WMT17), validate on previous years, and test on the current year (see Table~\ref{tab:wmt_data} for details). Our evaluations are conducted on official WMT22--24 datasets. WMT22~\cite{freitag-etal-2022-results}: Contains MQM and DA+SQM subsets with 16 language pairs, 40 systems, and 392,647 segments. WMT23~\cite{freitag-etal-2023-results}: Includes 282,926 segments over 11 language pairs and 29 MT systems. WMT24~\cite{freitag-etal-2024-llms}: For the high-quality MQM subset, there are 3 language pairs, 32 systems, and 68,502 segments; the ESA subset includes 232,289 segments covering 9 language pairs and 40 systems.

\begin{table*}[h!]
\centering
\def\arraystretch{1.1}% 增加行间距
\setlength{\tabcolsep}{4pt}% 增加列间距
\resizebox{\linewidth}{!}{%
\begin{tabular}{lccccc}
\toprule
\textbf{Train set} & 
\textbf{Val set} & 
\textbf{Benchmark/Test set}& 
\textbf{\#Languages in Test set} &
\textbf{\#Segments in Test set} &
\textbf{Subsets in Test set} \\
\midrule

WMT17-20 & WMT21 & WMT22 & 16 language pairs & 392,647 segments & MQM, DA \\
WMT17-21 & WMT22 & WMT23 & 11 language pairs & 282,926 segments & MQM, DA+SQM \\
WMT17-22 & WMT23 & WMT24 & 12 language pairs & 232,289 segments & MQM, ESA \\

\bottomrule 
\end{tabular}%
}
\caption{WMT22-24 Benchmark Descriptions.}
\label{tab:wmt_data}
\end{table*}

\subsubsection{ACES Score}\label{appendix:aces}

In this paper, we follow the original ACES Score calculation~\cite{amrhein-etal-2022-aces,moghe2025machine}, which provides a comprehensive assessment by combining performance on various error types with appropriate weightings. As shown in Equation~\ref{ACES_EQ}, the ACES Score assigns higher weights (5) to critical error categories such as addition, omission, mistranslation, overtranslation, and undertranslation, while giving lower weights to categories like untranslated segments (1), wrong language (1), and punctuation errors (0.1). 

\begin{equation}
\text{ACES} = sum \left\{
\begin{array}{c}
5 * \tau_{\text{addition}} \\
5 * \tau_{\text{omission}} \\
5 * \tau_{\text{mistranslation}} \\
1 * \tau_{\text{untranslated}} \\
1 * \tau_{\text{do not translate}} \\
5 * \tau_{\text{overtranslation}} \\
5 * \tau_{\text{undertranslation}} \\
1 * \tau_{\text{real-world knowledge}} \\
1 * \tau_{\text{wrong language}} \\
0.1 * \tau_{\text{punctuation}}
\end{array}
\right\}
\label{ACES_EQ}
\end{equation}

This weighting scheme reflects the relative impact of different error types on overall translation quality. For more details on the ACES challenge set and the development of this scoring methodology, we refer readers to \citet{amrhein-etal-2022-aces} and \citet{moghe2025machine}.

\subsection{Pairwise Data Construction for Reward Modeling}\label{appendix:pairwise_data}
We construct pairwise preference training and validation data using the original raw human ratings for each translation. Specifically, given the same source and reference sentence pair $(\mathit{src},\mathit{ref})$, we examine human ratings for different translations and construct preference pairs $(\mathit{mt^+},\mathit{mt^-})$ where the human rating for $\mathit{h_{mt^+}}$ is higher than that for $\mathit{h_{mt^-}}$.

For DA (Direct Assessment) data with a [0,100] scale, we set a rating difference threshold of 25 points, following the common understanding that translations differing by less than 25 points should be considered of equivalent quality. 

For MQM (Multidimensional Quality Metrics) data with a [0,25] scale, we use a much smaller threshold of 0.1, as MQM annotations are more fine-grained, where even small differences like punctuation errors can meaningfully impact translation quality. 

Once we construct the pairwise preference data, we format inputs differently depending on the foundation model architecture (see Figure~\ref{fig:remedy_data_format} for more details).

Finally, we evaluate ReMedy with the official testset directly for each individual translation $(\mathit{src},\mathit{mt},\mathit{ref^*})$, without doing any data preprocessing steps. The final meta-evaluation is done by the official MTME tool.

\begin{figure*}[t]
\centering
\begin{minipage}{0.8\textwidth}
\label{appendix:pairwise_data}

\vspace{0.5em}
\noindent\textbf{Encoder-only models.} For encoder-only models, we use a simple concatenation format:

\begin{lstlisting}
# Preferred translation pair
    chosen = [                                        
        {src_lang}: {src}, {Reference}: {ref*}, {tgt_lang}: {mt+}
    ]
# Non-preferred translation pair
    rejected = [
        {src_lang}: {src}, {Reference}: {ref*}, {tgt_lang}: {mt-}
    ]   
\end{lstlisting}

\vspace{0.5em}
\noindent\textbf{Decoder-only models.} For decoder-only models, we use a chat template format with paired preferred and non-preferred examples:
\begin{lstlisting}
# Preferred translation pair
    chosen = [
        {'role': 'user', 
         'content': "Translate the following {src_lang} text into natural, 
                    fluent {tgt_lang} sentence while preserving the original 
                    meaning. You are also given a translation template.
                    {src_lang}:{src}
                    Template:{ref*}
                    {tgt_lang}:"},
        {'role': 'assistant', 'content': {mt+}}
    ]
# Non-preferred translation pair
    rejected = [
        {'role': 'user', 
         'content': "Translate the following {src_lang} text into natural, 
                    fluent {tgt_lang} sentence while preserving the original 
                    meaning. You are also given a translation template.
                    {src_lang}:{src}
                    Template:{ref*}
                    {tgt_lang}:"},
        {'role': 'assistant', 'content': {mt-}}
    ]
\end{lstlisting}

\noindent Where \{src\_lang\}, \{tgt\_lang\} represent source and target language, $\mathit{src}$, $\mathit{ref^*}$ denote the source and reference sentences, and $\mathit{mt^+}$ and $\mathit{mt^-}$ represent the preferred and non-preferred translations.
\end{minipage}
\caption{ReMedy data format for training and inference}
\label{fig:remedy_data_format}
\end{figure*}

\subsection{Additional Results}\label{appendix:results}

\subsubsection{WMT22}

We list the full results of WMT22 in Table~\ref{tab:wmt22_all}, demonstrating the performance of various metric systems on both MQM and DA subsets. Note that all closed models and Llama2-EAPrompt do not validate their results on the DA set.

\begin{table*}[h!]
\centering
\def\arraystretch{1.1}% 增加行间距
\setlength{\tabcolsep}{4pt}% 增加列间距
\resizebox{\linewidth}{!}{%
\begin{tabular}{llcccccc||ccc}
\toprule
\multirow{2}{*}{\textbf{Type}} & 
\multirow{2}{*}{\textbf{Methods}}& 
\multirow{2}{*}{$\theta$} & 
\multirow{2}{*}{\textbf{ref?}} &
\multicolumn{2}{c}{\textbf{System-Level Acc}} & \multicolumn{2}{c}{\textbf{Segment-Level Acc}} &
\multicolumn{3}{c}{\textbf{Avg Corr}}\\

\cmidrule(lr){5-6} \cmidrule(lr){7-8} \cmidrule(lr){9-11}
      & 
      & 
      & 
      & \textbf{MQM}
      & \textbf{DA}
      & \textbf{MQM}
      & \textbf{DA}
      & \textbf{MQM}
      & \textbf{DA}
      & \textbf{All}\\
 \midrule

\multirow{10}{*}{\begin{tabular}{c}\textbf{Closed}\\\textbf{Models}\end{tabular}} 

& GEMBA-ChatGPT (\textit{P}) & 175B & \cmark & 81.0\% & - & 50.1\% & - & 65.6\% & - & -\\

& GEMBA-GPT4 (\textit{P}) & - & \cmark & 89.8\% & - & 55.6\% & - & 72.7\% & - & -\\

& PaLM (\textit{P})     & 540B & \cmark & 90.1\% & - & 50.8\% & - & 70.5\% & - & -\\

& PaLM-2 BISON (\textit{R}) & <340B & \cmark & 88.0\% & - & 57.3\%  & - & 72.7\% & - & -\\
& PaLM-2 BISON (\textit{GC}) & <340B & \cmark & 86.1\% & - & 54.8\% & - & 70.5\% & - & -\\

& PaLM-2 UNICORN (\textit{R}) & \textasciitilde340B & \cmark & 87.6\% & - & 58.0\% & - & 72.8\% & - & -\\

& PaLM (\textit{P}) & 540B & \xmark & 84.3\% & - & 50.3\% & - & 67.3\% & - & -\\

& PaLM-2 BISON (\textit{R}) & - & \xmark & 87.6\% & - & 57.5\% & - & 72.6\% & - & -\\

& PaLM-2 BISON (\textit{GC}) & - & \xmark & 86.1\% & - &  53.2\% & - & 60.7\% & - & -\\

& PaLM-2 UNICORN (\textit{GC}) & - & \xmark & 86.1\% & - &  52.9\% & - & 69.5\% & - & -\\

\midrule

\multirow{10}{*}{\begin{tabular}{c}\textbf{Open}\\\textbf{Models}\end{tabular}}

& Llama2-EAPrompt (\textit{P}) & 70B & \cmark & 85.4\% & - & 52.3\% & - & 68.9\% & - & -\\

& COMET-22-DA (\textit{R}) & 0.5B & \cmark & 82.8\% & 86.4\% & 54.5\% & 55.4\% & 68.7\% & 70.9\% & 69.8\% \\

& COMET-22 (\textit{R}) & ensemble & \cmark & 83.9\% & 85.8\% & 57.3\% & \textbf{57.2\%} & 70.6\% & 71.5\% & 71.0\% \\

& MetricX-XXL (\textit{R}) & 13B & \cmark & 85.0\% & 86.5\% & 58.8\% & 55.6\% & 71.9\% & 71.1\% & 71.5\% \\

& Llama2-EAPrompt (\textit{P}) & 70B & \xmark & 85.8\% & - & 52.0\% & - & 68.9\% & - & - \\

& COMETKiwi (\textit{R}) & ensemble & \xmark & 78.8\% & 85.4\% & 55.5\% & \underline{56.5\%} & 67.2\% & 71.0\% & 69.1\% \\

\cmidrule(lr){2-10}
& \textnormal{ReMedy}$_\textnormal{xlm-r}$ \textnormal{(Ours)} & 0.5B & \cmark & 85.8\% & 86.6\% & 55.4\% & 55.6\% & 70.6\% & \underline{71.1\%} & 70.9\% \\

& \textnormal{ReMedy}$_\textnormal{gemma2}$ \textnormal{(Ours)} & 2B & \cmark & 90.5\% & 86.2\% & 55.9\% & 53.9\% & 73.2\% & 70.0\% & 71.6\% \\

%& \textnormal{ReMedy}_\textnormal{gemma2}\textnormal{-raw (Ours)} & 9B & \cmark & 91.2\% & 86.4\% & 58.1\% & 57.3\% & 74.7\% & 71.8\% & 73.3\% \\

%& \textnormal{ReMedy}_\textnormal{gemma2}\textnormal{-sigmoid (Ours)} & 9B & \cmark & 90.5\% & 87.3\% & 58.8\% & 57.5\% & 74.7\% & 72.4\% & 73.6\% \\

& \textnormal{ReMedy}$_\textnormal{gemma2}$\textnormal{(Ours)} & 9B & \cmark & \textbf{91.2\%} & \textbf{87.7\%} & \textbf{58.9\%} & 56.0\% & \textbf{75.1\%} & \textbf{71.9\%} & \textbf{73.5\%} \\

& \textnormal{ReMedy}$_\textnormal{gemma2}$ \textnormal{(Ours)} & 9B & \xmark & \underline{89.4\%} & \underline{85.8\%} & \underline{57.8\%} & 54.3\% & \underline{73.6\%} & 70.0\% & \underline{71.8\%} \\

\bottomrule 
\end{tabular}%
}
\caption{Evaluation on WMT22 MQM and DA set. The system-level results are Pairwise Accuracy proposed by \citet{kocmi2021ship}, and segment-level results are based on the group-by-item pairwise accuracy with tie calibration~\cite{deutsch2023ties}. P denotes prompting (no tuning); R and GC represent Regression and Generative Classification training objectives. \textbf{Bold} and \underline{underline} indicate the best metric and QE (no reference) models. COMET-22 and COMETKiwi are ensembled with 5x and 6x 0.5B models, respectively.}
\label{tab:wmt22_all}
\end{table*}

\subsubsection{WMT24}

For WMT24, we present the full results for both MQM and ESA subsets in Table~\ref{tab:wmt24_all}. Following the WMT24 official meta evaluation protocol~\cite{freitag-etal-2024-llms}, we use the MQM subset for our primary comparisons as it provides higher-quality human annotations than the crowd-sourced ESA set. Our analysis reveals that ReMedy-9B performs slightly worse on the ESA subset, primarily due to lower performance on English-Hindi and English-Icelandic language pairs (complete evaluation results available in our repository\footnote{\url{https://github.com/Smu-Tan/Remedy}}). 

This underperformance likely stems from these languages being relatively low-resource in the pre-trained Gemma2 model. Interestingly, ReMedy-9B-22 still outperforms MetricX and COMET on previously unseen extremely low-resource language pairs such as English-Livonian and Yakut-Russian in the WMT22 test set. We intend to investigate these performance differences in future work.

\begin{table*}[h!]
\centering
\def\arraystretch{1.0}% 增加行间距
\setlength{\tabcolsep}{8pt}% 增加列间距
\resizebox{\linewidth}{!}{%
\begin{tabular}{lccccc||cc}
\toprule
\multirow{2}{*}{\textbf{Methods}}& 
\multirow{2}{*}{$\theta$} & 
\multicolumn{2}{c}{\textbf{System-Level SPA}} & \multicolumn{2}{c}{\textbf{Segment-Level Acc}} &
\multicolumn{2}{c}{\textbf{Avg Corr}}\\

\cmidrule(lr){3-4} \cmidrule(lr){5-6} \cmidrule(lr){7-8}
      & 
      & \textbf{MQM}
      & \textbf{DA}
      & \textbf{MQM}
      & \textbf{DA}
      & \textbf{MQM}
      & \textbf{DA} \\
      
\midrule

GEMBA-ESA (\textit{P}) & - & 85.2\% & 81.5\% & 57.6\% & 42.2\% & 71.4\% & 61.8\% \\

XCOMET (\textit{R}) & 24B & \textbf{86.6\%} & 85.4\% & 57.6\%  & 56.3\% & 72.1\% & 70.9\%\\

MetricX-24-Hybrid (\textit{R}) & 13B & 85.6\% & \textbf{86.4\%} & 58.5\% & \textbf{56.5\%} & 72.1\% & \textbf{71.4\%} \\

\midrule

ReMedy$_\textnormal{Gemma2}$ \textnormal{(Ours)} & 9B & 86.4\%	& 83.0\% & \textbf{60.2\%} & 55.7\% & \textbf{73.3\%} & 69.3\%\\

\bottomrule 
\end{tabular}%
}
\caption{Evaluation on WMT24 MQM and DA set. }
\label{tab:wmt24_all}
\end{table*}

\subsection{Reward Calibration Analysis}\label{appendix:reward_calibration}

In this section, we demonstrate how our entropy-guided temperature selection adapts to different reward distributions, maximizing the information content of the final calibrated scores. By selecting the temperature that maximizes Shannon entropy across 20 uniform bins in the [0,1] interval, we ensure calibrated scores utilize the full range effectively, preventing clustering and preserving meaningful distinctions between translations of varying quality. Our entropy maximization can be formulated below in Eq ~\ref{equation:entropy}:

\begin{equation}\label{equation:entropy}
    \tau^* = \underset{\tau}{\arg\max} \, H(P_\tau) = \underset{\tau}{\arg\max} \, -\sum_{i=1}^{20} p_i^\tau \log p_i^\tau
\end{equation}

where $P_\tau = \{p_1^\tau, p_2^\tau, ..., p_{20}^\tau\}$ represents the distribution of calibrated scores across 20 bins when using temperature $\tau$. This approach dynamically adapts to different reward distributions, providing optimal discrimination where it matters most.

\subsubsection{High Temperature Case Study: Right-Skewed Distributions}

\begin{figure*}[h!]
    \centering
    \includegraphics[width=0.8\linewidth]{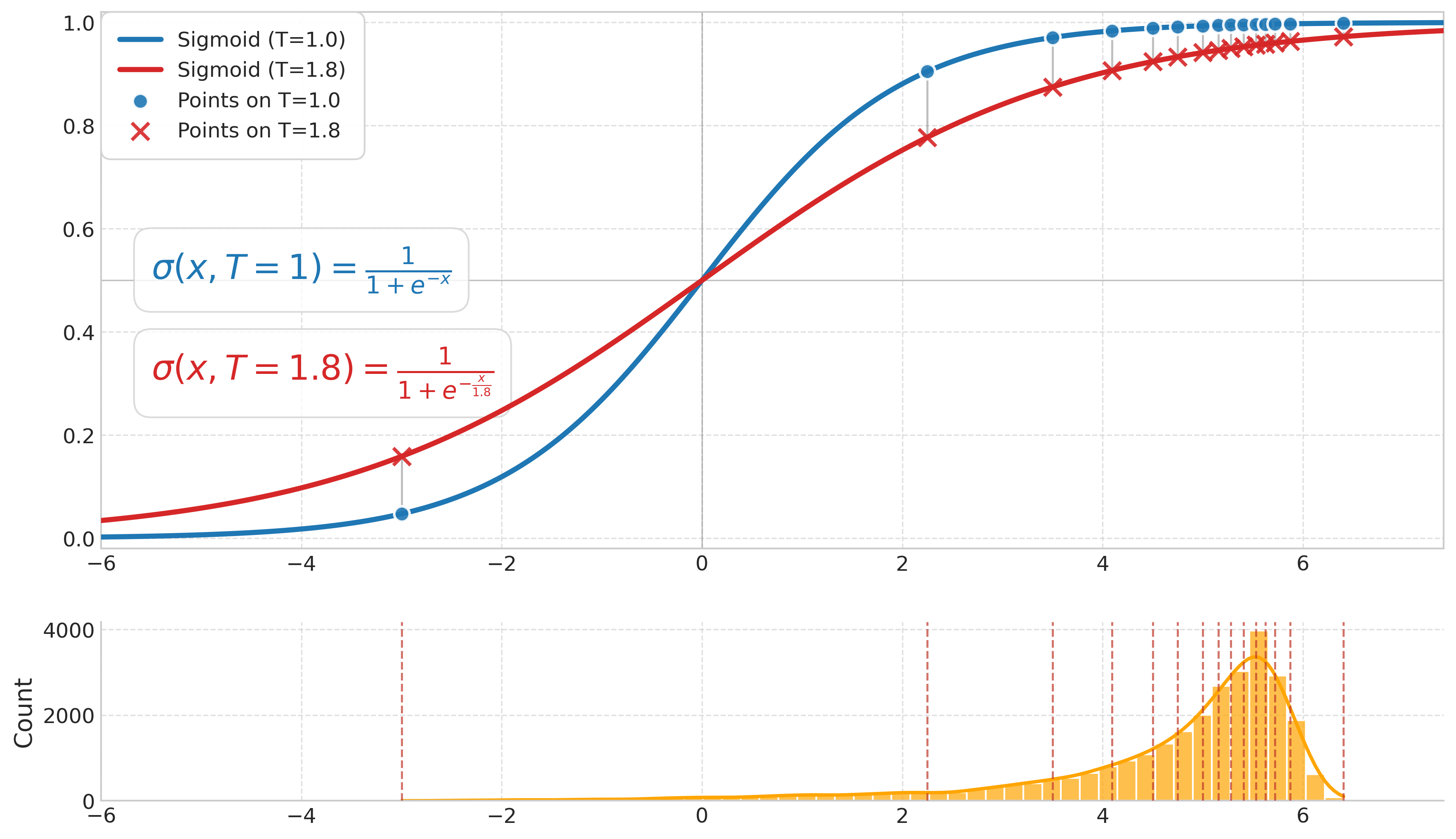}
    \caption{Reward calibration with high temperature}
    \label{fig:high_temp}
    %\vspace{-4mm}
\end{figure*}

Figure~\ref{fig:high_temp} illustrates our entropy-guided reward calibration for WMT22 English-German translation submissions. For high-quality MT systems, raw rewards are typically concentrated in the upper range, creating a right-skewed distribution.

The top panel shows two sigmoid functions with different temperature values: the standard sigmoid with $T=1.0$ (blue) and our entropy-optimized sigmoid with $T=1.8$ (red). The mathematical formulations display how the temperature parameter affects the steepness of the curve. The bottom panel shows the histogram of raw reward values from ReMedy-9B, where rewards are heavily concentrated between 4 and 6, reflecting the high quality of WMT22 English-German translation submissions.

With a standard sigmoid ($T=1.0$), most high reward values would be mapped to scores very close to 1.0, making distinguishing between good and excellent translations difficult. By increasing the temperature to $T=1.8$, the sigmoid curve is horizontally stretched, creating more separation between high-quality translations in the final [0,1] score range. The vertical dashed red lines illustrate how specific histogram bins map to points on the sigmoid curve.

Table~\ref{tab:sigmoid_comparison_high} shows numerically how the increased temperature creates meaningful separation between high-quality translations. For example, raw scores of 4.09 and 5.00 would receive nearly identical scores (0.984 vs. 0.993) with the standard sigmoid, but more distinguishable scores (0.907 vs. 0.941) with our calibrated approach.

\begin{table}[h!]
    \centering
    \begin{tabular}{ccc}
        \hline
        Raw score ($x$) & $\sigma(x, T=1.0)$ & $\sigma(x, T=1.8)$  \\
        \hline
        -3.00000 & 0.04743 & 0.15887 \\
        2.25000 & 0.90465 & 0.77730 \\
        3.50000 & 0.97069 & 0.87484 \\
        4.09375 & 0.98360 & 0.90673 \\
        4.50000 & 0.98901 & 0.92414 \\
        4.75000 & 0.99142 & 0.93332 \\
        5.00000 & 0.99331 & 0.94146 \\
        5.15625 & 0.99427 & 0.94607 \\
        5.28125 & 0.99494 & 0.94950 \\
        5.40625 & 0.99553 & 0.95273 \\
        5.53125 & 0.99605 & 0.95576 \\
        5.62500 & 0.99641 & 0.95791 \\
        5.71875 & 0.99673 & 0.95996 \\
        5.87500 & 0.99720 & 0.96317 \\
        6.40625 & 0.99835 & 0.97232 \\
        \hline
    \end{tabular}
\caption{Sigmoid calibration values for right-skewed reward distributions of high-quality submission systems. The higher temperature ($T=1.8$) creates larger separation between high rewards that would otherwise cluster near 1.0 with the standard sigmoid. This enables better discrimination between good and no-error translations. These scores correspond to values in Figure~\ref{fig:high_temp}.}
\label{tab:sigmoid_comparison_high}
\end{table}

\subsubsection{Low-Normal Temperature Case Study: Rewards with Normally Distribution}

\begin{figure*}[h!]
    \centering
    \includegraphics[width=0.8\linewidth]{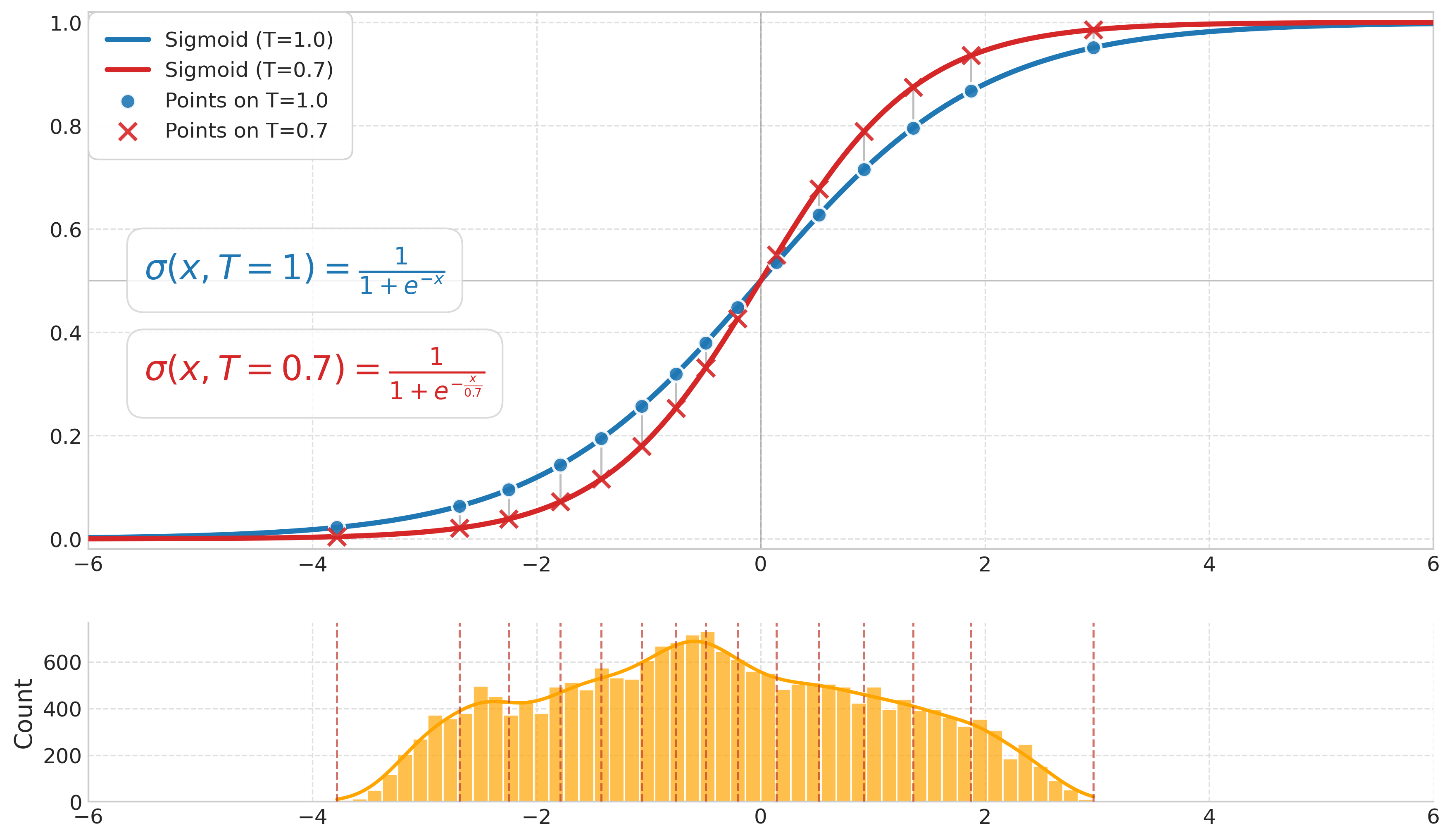}
    \caption{Reward calibration with low temperature}
    \label{fig:low_temp}
    %\vspace{-4mm}
\end{figure*}

On the other hand, Figure~\ref{fig:low_temp} demonstrates calibration for a more evenly distributed set of raw rewards (approximately normally distributed around 0). Such distribution appears when evaluating diverse MT systems with varying quality levels. Here, our entropy-guided approach selects a temperature of $T=0.7$, lower than the standard $T=1.0$.

With $T=0.7$, the sigmoid curve is more compressed, making it steeper around the center. This compression provides enhanced discrimination for translations in the mid-quality range, where most reward values are concentrated in this distribution. The histogram in the bottom panel confirms the balanced distribution of raw rewards, and the vertical dashed lines illustrate the mapping between histogram bins and sigmoid values.

Table~\ref{tab:sigmoid_comparison_normal} demonstrates how the lower temperature creates greater separation in the central region of the distribution. For instance, raw scores of -0.76 and 0.52 show larger differences with $T=0.7$ (0.253 vs. 0.677) compared to $T=1.0$ (0.319 vs. 0.627), improving our ability to discriminate between average-quality translations.

\begin{table}[h!]
    \centering
    \begin{tabular}{ccc}
        \hline
        Raw score ($x$) & $\sigma(x, T=1.0)$ & $\sigma(x, T=0.7)$ \\
        \hline
        -3.78125 & 0.02229 & 0.00449  \\
        -2.68750 & 0.06371 & 0.02106  \\
        -2.25000 & 0.09535 & 0.03863  \\
        -1.78906 & 0.14319 & 0.07204  \\
        -1.42188 & 0.19437 & 0.11596  \\
        -1.06250 & 0.25683 & 0.17978  \\
        -0.75781 & 0.31912 & 0.25302  \\
        -0.49219 & 0.37938 & 0.33112  \\
        -0.20703 & 0.44843 & 0.42659  \\
        0.14062 & 0.53510 & 0.55005   \\
        0.51953 & 0.62704 & 0.67747   \\
        0.92188 & 0.71542 & 0.78868   \\
        1.35938 & 0.79566 & 0.87457   \\
        1.87500 & 0.86704 & 0.93575   \\
        2.96875 & 0.95114 & 0.98581   \\
        \hline
    \end{tabular}
\caption{Sigmoid calibration values for evenly distributed rewards. The lower temperature ($T=0.7$) creates larger separation in the central region where most scores are concentrated, improving discrimination between translations of moderate quality. These scores correspond to values in Figure~\ref{fig:low_temp}.}
\label{tab:sigmoid_comparison_normal}
\end{table}

These case studies demonstrate how our entropy-guided temperature selection dynamically adapts to different reward distributions. This approach is proved to yield better alignment with human judgments (see Table~\ref{tab:ablations}) when evaluating diverse MT systems that may produce translations clustered in different quality ranges, ensuring optimal discrimination across the entire quality spectrum.

\end{document}